\pdfoutput=1
\documentclass{article}

\usepackage{natbib}
\usepackage[final]{nips_2017}

\usepackage[utf8]{inputenc} 
\usepackage[T1]{fontenc}    
\usepackage{hyperref}       
\usepackage{url}            
\usepackage{booktabs}       
\usepackage{amsfonts}       
\usepackage{amsmath}
\usepackage{nicefrac}       
\usepackage{microtype}      
\usepackage{graphicx}
\newcommand*\diff{\mathop{}\!\mathrm{d}}

\title{Deep Learning for Patient-Specific Kidney Graft Survival Analysis}

\author{
  Margaux Luck\thanks{Equal contribution} \\
  University Paris Descartes and Institut Hypercube\\
  Paris, France \\
  \texttt{margaux.luck@gmail.com}  \\
  \And
  Tristan Sylvain\footnotemark[1] \\
  Montreal Institute for Learning Algorithms \\
  Montreal, Quebec, Canada \\
  \texttt{tristan.sylvain@gmail.com}
  \AND
  H\'elo\"ise Cardinal \\
  Centre de Recherche and Department of Medicine, Nephrology, \\
  Centre Hospitalier de l’Universit\'e de Montr\'eal \\
  Montreal, Quebec, Canada \\
  \texttt{heloise.cardinal.chum@ssss.gouv.qc.ca} \\
   \And
  Andrea Lodi \\
  Canada Excellence Research Chair \\
  École Polytechnique de Montréal \\
  Montreal, Quebec, Canada \\
  \texttt{andrea.lodi@polymtl.ca} \\
   \And
   Yoshua Bengio \\
  Montreal Institute for Learning Algorithms \\
  Montreal, Quebec, Canada \\
  \texttt{yoshua.bengio@umontreal.ca}
}

\begin{document}
\maketitle

\begin{abstract}
	An accurate model of patient-specific kidney graft survival distributions can help to improve shared-decision making in the treatment and care of patients. In this paper, we propose a deep learning method that directly models the survival function instead of estimating the hazard function to predict survival times for graft patients based on the principle of multi-task learning. By learning to jointly predict the time of the event, and its rank in the cox partial log likelihood framework, our deep learning approach outperforms, in terms of survival time prediction quality and concordance index, other common methods for survival analysis, including the Cox Proportional Hazards model and a network trained on the cox partial log-likelihood.
\end{abstract}

\section{Introduction}
Survival data are characterized by the fact that we do not know many of the outcome values (e.g. death or disease recurrence in medical studies) because the event might not have occurred within the fixed period of the study or because subjects could move out of town or decide to drop out at any time. Instead, the date of the last visit (censoring time) provides a lower bound on the survival time, defined as the length of time between two events, called intake and endpoint. Such datasets are considered {\em censored}. The Cox proportional hazards model and other parametric survival distributions such as Weibull, exponential and log-normal, have been widely used for the analysis of the survival time of a population by researchers and clinicians to test for significant risk factors affecting survival.

These models however suffer from several limitations when the objective is to get a precise estimation of the survival time of an individual patient. Chun-Nam Yu et al.~\cite{yu2011learning} lists two main issues of such models. The distributions chosen to model the data make it hard to provide accurate predictions in terms of survival time. Cox models use the Proportional Hazard Assumption (PHA) that constrains the model in a way that the effect of a given feature does not vary over time. This is a restrictive condition in practice.

Some models, such as the stratified Cox model, attempt to deal with the limitations of the PHA. Chun-Nam Yu et al.~\cite{yu2011learning} also deals with both problems by introducing the multi-task logistic regressions (MTLR) model, which introduces multiple logistic regression to model survival after a given time. In addition, there have been publications in the past comparing machine learning methods such as neural networks~\cite{hashemian2013comparison, jerez2003combined} with linear Cox models. Blaz Zupen et al.~\cite{zupan2000machine} also proposed a framework that enables the use of machine learning methods such as Bayes classifiers and decision trees for survival analysis.

In this paper we introduce a novel deep learning method for survival analysis, based on multi-task learning. We jointly predict the time of the event, and its rank in the cox partial log-likelihood framework. This allows better generalization, due in part to the fact that the model is able to account for the temporal aspect of the predictions. We compute the concordance index metric \cite{doi:10.1001/jama.1982.03320430047030} (C-index in what follows) in order to compare our results with the widely adopted Cox model. We demonstrate the performance of our method experimentally on real-life survival datasets where it yields better results in terms of C-index than the previous state-of-the-art methods.

\section{Background: Survival time analysis}
\subsection{Notations}
In contrast to most common regression problems, survival data analysis has three main characteristics: (1) it examines the relationships of survival distributions to features; (2) it models the time it takes for events to occur, and (3) the event we want to predict (such as time of death) is not always observed. Sometimes, a patient will drop out of the study (i.e., voluntarily or because he was still alive at the end of the study). We call such datasets right-censored.   

Let $T$ be a continuous random variable representing survival time. The survival function $S(t)$ is the probability of a patient surviving longer than $t$, i.e., 
\begin{align*}
S(t) = P(T \geq t).
\end{align*}

The hazard function denoted by $\lambda (t)$ is the instant probability that the event occurs knowing that it did not occur before $t$.
We can define $\lambda (t)$ as
\[\lambda (t) = \lim_{dt \rightarrow 0} \frac{P(t \leq T < t + dt | T \geq t)}{dt}.\]

The survival function can be expressed as a function of the hazard at all durations up to $t$
\[ S(t) = \exp\Big(-\int_0 ^t \lambda(x) \diff x\Big).\]
As medical events are granular by nature, a given \emph{unique time} (in a given time unit, such as a month for our dataset) can correspond to multiple events (such as having 10 patients reject their graft on January). Such events are \emph{tied}, making comparisons more complex, and requiring a modification to the loss function.

\subsection{Linear models: Cox proportional hazards model}

Some common approaches attempt to model the hazard function using the proportional hazards assumption.
Different modelizations of $\lambda$ have been considered. Among the most well known, the semi-parametric Cox proportional hazards model~\cite{cox1972regression} defines $\lambda$ at time $t$ for an individual with features $\mathbf{x_{i}}$ as
\[ \lambda_{i} (t | \mathbf{x_{i}}) = \lambda_0(t) \exp (\theta \cdot \mathbf{x_{i}}). \]

The Cox proportional hazards model can be viewed as consisting of two clearly separate parts: (1) the underlying baseline hazard function $\lambda_0(t)$, describing how the risk of event per time unit changes over time at baseline levels of features, and (2) the effect parameters $\exp (\theta \cdot \mathbf{x_{i}})$, describing how the hazard varies in response to explanatory features.

Note that the Cox proportional hazards model is not commonly used in the literature to perform prediction on new cases, but rather to characterize disease progression on existing cases, by highlighting the importance of the different features~\cite{rao2009comprehensive}. 
As the baseline hazard function is never directly estimated, computing survival predictions is not directly possible without additional assumptions.

\subsection{Non-linear models}
Other approaches have been applied to survival data.
Random Survival Forests~\cite{ishwaran2008random} are based on an ensemble of trees to estimate the cumulative hazard function. This extends Breiman's Random Forests method to take right-censored survival data into account.

Deep survival ~\cite{katzman2016deep} is an extension of the Cox model that uses a deep neural network to parametrize the hazard function. The part of our model that is trained on a Cox partial likelihood loss differs from their approach in the following regards:
\begin{itemize}
\item We perform the optimization per-batch instead of on the full dataset.
\item We adapt the loss function to account for ties, using Efron's approximation.
\item We did not limit ourselves to modeling the logarithm of the hazard function, and tried other forms. Modeling the hazard function directly yielded an improvement.
\end{itemize}

Ranganath et al. \cite{ranganath2016deep} uses deep exponential families (i.e., a class of latent variable models inspired by the hidden structures used in deep neural networks) to model event time.

\subsection{Standard evaluation: the concordance index}

To compare our models we use the C-index~\cite{doi:10.1001/jama.1982.03320430047030}. The C-index is a standard measure in survival analysis that estimates how good the model is at ranking survival times by calculating the probability of correctly ranking the event time of cases taken two at a time. It can be seen as a generalization of the Area Under the Receiver Operating Characteristic Curve (AUROC) to regression problems and thus can handle right-censored data~\cite{steck2008ranking}. Let $T_i$ denote the survival time for individual $i$ and $E_i$ be the associated event, censored or uncensored. The $(T_{i}, E_{i}), ..., (T_{n}, E_{N})$ are all the events in the dataset. Considering all possible pairs $(T_{i}, E_{i}),(T_{j}, E_{j})$ for $i \leq j$, the C-index is calculated by considering the number of pairs correctly ordered by the model divided by the total number of admissible pairs. For our particular case of right censoring, a pair is considered admissible if it can be ordered in a meaningful way. A pair cannot be ordered if the events are both right-censored or if the earliest time in the pair is censored. A tied pair is counted as half correct in accordance with standard implementations of the C-index. Finally, a C-index equal to 1 indicates perfect prediction whereas a C-index equal to 0.5 indicates a random prediction. Survival models typically yield a C-index between 0.6 and 0.7.

\section{A deep learning survival model}
In this section we describe our main contribution. As opposed to most previous methods that attempt to estimate the survival or hazard function, we construct a deep neural network model to directly compute the time of the event (here, graft failure). Our model attempts to predict the probability of being alive at time $t$, for $t \in [0, T]$. The chosen modeling task is related to the problem of ordinal regression, with the exception that we have to take into account censored events. 

\subsection{Model}
\begin{figure}[!t]
\centering
\includegraphics[scale=.7]{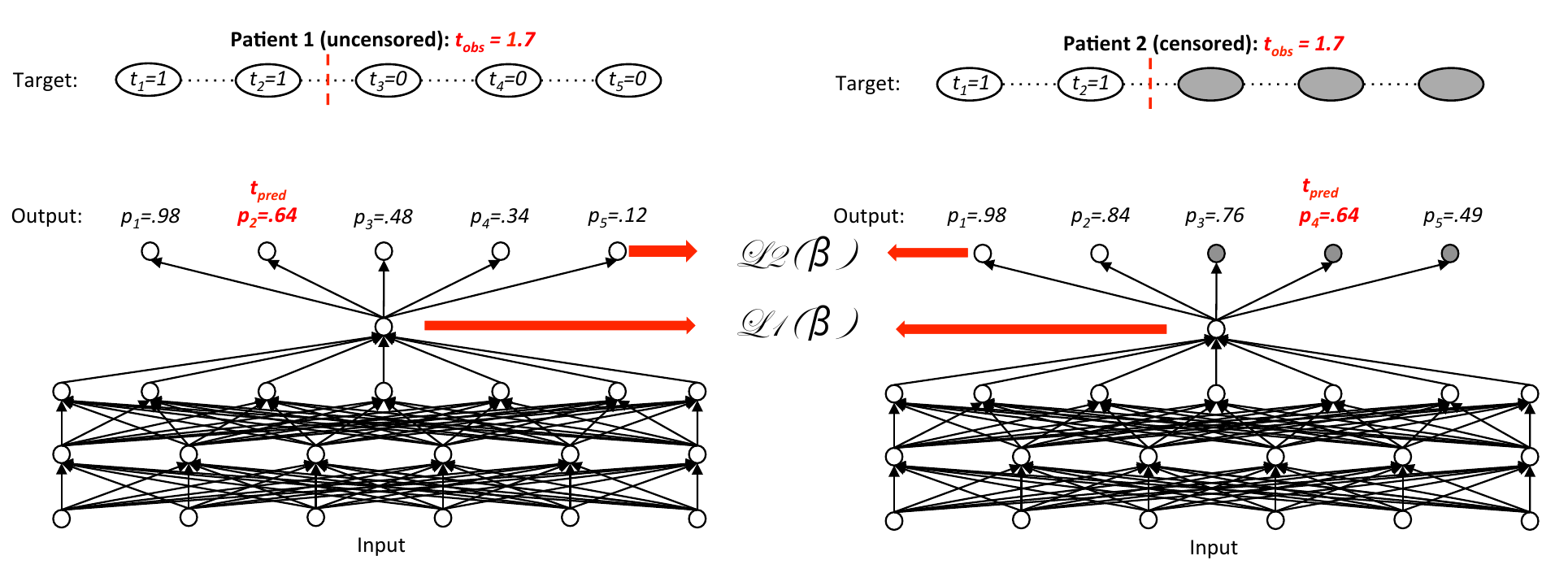}
\caption{Our deep learning model. (Left) The case of an uncensored patient. (Right) The case of a censored patient. The greyed out output units in the censored case correspond to censored units for which no loss is computed and no error back-propagation is done. Two losses are computed: a ranking loss on the bottleneck scalar unit and a cumulative time prediction on the top output units.}
\label{fig:model}
\end{figure}

The proposed model takes as input the different continuous and discrete features characterizing a patient and, in the case of the main dataset of our paper, a donor-recipient couple. The second-to-last layer consists of a single unit with linear activation. The value outputted, denoted $s^{(1)}$ in what follows is used to estimate the hazard function and can be considered a score indicative of the time of graft failure, and thus compare two patients. This allows us to compute the first loss, the Cox partial log-likelihood. The final layer has $T$ units, where $T$ is the number of time units (in our case, years or months) considered in the study. The output is denoted $s^{(2)}$ in what follows. We use sigmoid activations for the units of the final layer. The output value at index $t$ corresponds to the probability of not experiencing graft failure at time $t$ (in our case at the $t$-th year). The second loss penalizes wrongly predicted times in a manner consistent with losses used in isotonic regression. The model is shown in a simplified form (removing some layers for clarity) in Figure~\ref{fig:model}. The two cases (censored and non-censored) are shown in the figure, which illustrates that when some right part of the history is missing, no loss and no gradient is computed for those output units.

\subsection{Handling ties and censored data}
The cost function we used is specifically adapted to the presence of ties and the censored nature of the dataset. It combines two losses.
The first one is Cox's partial likelihood using Efron's approximation to handle ties \cite{efron1977efficiency}. Let $Y_{i}$ be the observed time (either censoring or event time) of subject $i$. $C_{i}$ is $1$ if $t_i$ corresponds to an actual (non-censored) event, and $0$ if the event is censored. 
The loss is defined as follows:
\begin{equation}
\ell_{1} (s^{(1)}) = \sum_{j} \left( \sum_{i \in H_{j}} \log s^{(1)}_i - \sum_{\ell = 0}^{m-1} \log \left( \sum_{i:Y_{i} \geq t_{j}} s^{(1)}_i - \frac{l}{m} \sum_{i \in H_{j}} s^{(1)}_i \right) \right)
\end{equation}
where $t_{j}$ denote the unique times, $H_{j}$ the set of indices $i$ such that $Y_{i} = t_{j}$ and $C_{i}=1$ and $n_{j}=|H_{j}|$. Note that in the case of the Cox model, $s^{(1)} = e^{\theta \cdot X}$.

The second loss is adapted from the proposal of Menon et al. ~\cite{menon2012predicting} of combining a ranking loss with an isotonic regression. It is modified to take into account censored data. It is defined as
\begin{equation}
\ell_{2} (w) = \sum_{\text{Acc}(i, j)} \ell \left(\hat{s}^{(2)} \left(x_{j} ; w \right) - \hat{s}^{(2)}\left( x_{i};w \right), 1 \right) y_{i}(1-y_{j}) 
\end{equation}
where $\text{Acc}(i, j)$ selects acceptable pairs, that is: $i$ not censored, and at time of $i$'s event, $j$ is not censored; $\ell (\cdot,\cdot)$ is some convex loss function, here the $\ell_2$ distance and $\hat{s}$ is the nonlinear scoring function of the deep learning model.

\subsection{Evaluation procedure}
We first split the dataset into training (80\%) and test (20\%) sets in which the percentage of uncensored patients, and the proportion of events occurring per time-step is preserved. We performed hyper-parameter selection and early stopping on a subset of the training set (validation set corresponding to 20\% of the total dataset and having the same proportion of uncensored patients and events occurring per time-step).

As performance scores we used two measures: the C-index and the AUROC. Moreover, the C-index metric is the most common evaluation metric in the literature and as we said is a generalization of the AUROC to regression problems in the case of right-censored data. This has the advantage of allowing us to compare our method with other commonly used algorithms that do not necessarily output a meaningful predicted survival time, as is the case for the Cox proportional hazards model. Indeed, these two metrics can be seen as an evaluation of the pairwise ranking performance being the probability that a randomly drown superior/positive time example has a higher score than a randomly drawn inferior/negative example.

For the AUROC curve we slightly modify the original version in order to take into account the fact that the data are right-censored, namely
\[ A\left( \hat{s} \left( \cdot \right) \right) = Pr_{\text{Acc} \left[ \left( x_{i}, y_{i} \right) \left( x_{j}, y_{j} \right) \right] } \left[ \hat{s}^{(1)} \left( x_{i} \right) \geq \hat{s}^{(1)} \left( x_{j} \right) | y_{i} = 1, y_{j} = 0 \right] \]
where $\text{Acc} \left[ \left( x_{i}, y_{i} \right) \left( x_{j}, y_{j} \right) \right]$ is an acceptable pair, that is: $i$ is not censored, and at time event $i$ occurs, $j$ is not censored.

\section{Experiments}
\subsection{Dataset}
This study used data from the Scientific Registry of Transplant Recipients (SRTR). The SRTR data system includes data on all donor, wait-listed candidates, and transplant recipients in the US, submitted by the members of the Organ Procurement and Transplantation Network (OPTN). The Health Resources and Services Administration (HRSA), U.S. Department of Health and Human Services provides oversight to the activities of the OPTN and SRTR contractors. The dataset we used includes 131,709 deceased donor transplants between January 1, 2000 and December 31, 2014.  Recipients aged less than 18 years old and simultaneous multi-organ transplant recipients were excluded. The outcome of interest was graft failure, defined as return to dialysis, re-transplant, or death. 86104 patients (around 65\%) were censored. We used all available clinical and biological features characterizing the donor-recipient couple and transplant factors features potentially associated with graft failure rates, removing those that had more than 20 percent of missing values, 

\subsection{Preprocessing}
We completed missing values by replacing values for each feature by the median value for continuous features, and by the most common occurrence for categorical features. We chose to use this method because the number of missing values was very low (around 5 percent on average and up to 20 percent).
We used a one-hot encoding for categorical features, and unit scaling for continuous features. This resulted in a total of 436 features, all between $0$ and $1$.

\subsection{Hyper-parameters and training}
We used the Adam optimizer~\cite{kingma2014adam} with a learning rate of \textbf{$10^{-5}$}, and a batch size of \textbf{$32$}.
We used dropout, batch-normalization, $L_{1}$ and $L_{2}$ regularization and gradient clipping during training. The dropout rate for the different layers was optimized along with the other hyper-parameters.

Hyper-parameters were chosen through random search. The batch size had a large impact on the partial log-likelihood part of the loss, which is to be expected as it is a function of all pairwise combinations of elements in the batch, and therefore gains a lot of information from additional examples in the batch.

\subsection{Survival rate prediction}
We first aimed at evaluating the ranking of the patients in general using the C-index.  
We obtained a C-index of 0.655 which is higher than the C-index state of the art we obtained with the traditional Cox model using Efron's method for the loss on this dataset 0.65. Table \ref{cindex_different_models} summarizes the results.

\begin{table}[!ht]
\centering
\small
\caption{
{\bf C-index obtained for the different models tested.}}
\label{cindex_different_models}
\begin{tabular}{ |l|c|c|c|c|c| } 

\hline
Datasets & Cox Efron's & MLP Efron's & MLP rank & MLP Efron's + rank \\
\hline
SRTR & 0. 6504 & 0.6535 & 0.6302 & \textbf{0.6550}\\
\hline
\end{tabular}
\label{c_index_results}
\end{table}

An important point is that our algorithm allows us to use the features without applying any preprocessing~\cite{lecun2015deep}. Indeed, deep learning models are usually able to detect highly nonlinear effects of features which is not the case with the cox model.  For example, in the paper used as reference for the SRTR dataset, researchers required extensive effort and domain expertise to design features that are suitable for the Cox model~\cite{rao2009comprehensive}.
Moreover, the results we obtain show that learning on multiple objectives improves generalization on our specific task.

\begin{figure}[!t]
\centering
\includegraphics[scale=.55]{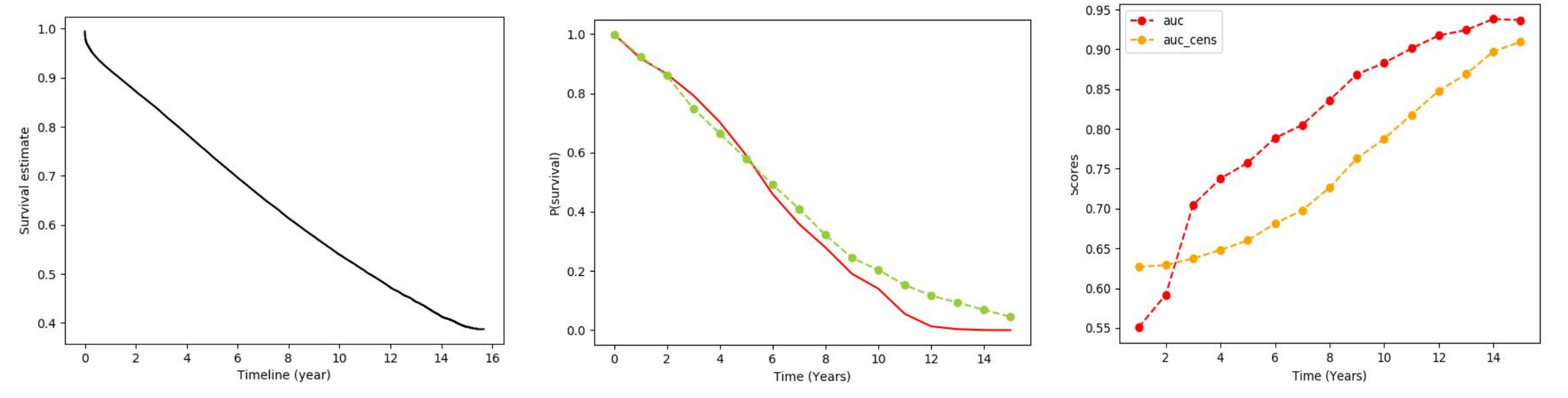}
\caption{(Left) Kaplan Meier curve (Middle) Median survival curve (dotted: averaged over the population, full: single individual)  (Right) AUROC (auc) and truncated AUROC (auc\_cens) at the different time thresholds predicted by our model}
\label{accuracy-roc}
\end{figure}

We also aimed to evaluate the predicted survival probabilities at different time thresholds. We focused on AUROC at each time threshold. \\
The models we trained had a certain trade-off between optimizing the two losses. As a result, achieving state of the art C-index came at the cost of slightly reduced ability to predict survival distributions per year. We focused on a choice of hyperparameters that slightly favored the latter in the plots that follow.\\
AUROC allows us to measure the quality of the prediction at each time step. Figure~\ref{accuracy-roc} (Right) shows the AUROC on the predicted probabilities for each observed year in the dataset. Our model performs worse predictions on the years close to the graft than on those that are distant. This can be explained by the fact that few patients die shortly after the transplant. It may be worthwhile to do data augmentation to increase the number of events close to the graft. However, this is difficult due to the high sparsity of the input matrix. \\

\subsection{Visualization}
\subsection{Survival rate prediction}

The Kaplan-Meier (KM) survival curve, shown in Figure~\ref{accuracy-roc} (Left) is a plot of the estimated survival probability against time~\cite{kaplan1958j}. In the KM curve the survival probability $S(t)$ is estimated as a step function where the value at time $t_{i}$ is calculated as follows: $S(t_{i}) = S(t_{i-1})(1-d_{i}/n_{i})$ with $d_{i}$, the number of events at $t_{i}$ and $n_{i}$ the number of patients alive just before $t_{i}$.

\subsection{Determining the influence of features}

\begin{figure}[!t]
\centering
\includegraphics[scale=.7]{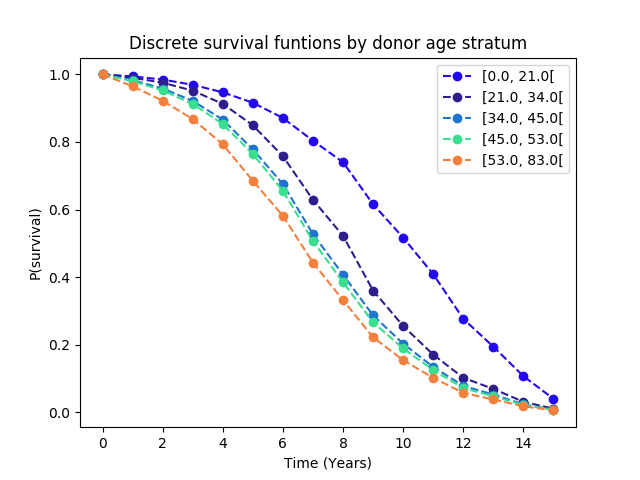}
\caption{Mean survival rate for different slice of donor age.}
\label{strata-donorage}
\end{figure}

As information on the role and importance of features is both important to clinicians, and hard to determine from a trained neural network, we performed a separate analysis on those points. One way to do so is to plot different survival curves for different strata of a given feature. Figure~\ref{strata-donorage} shows the mean survival curve obtained for different slices of donor age. According to the plot, we can deduce that the age of the donor seems to influence nonlinearly the graft survival prediction. In particular, donors aged less than 21 will lead to a significantly higher survival rate.

\subsection{Variable importance (VIMP)}
As has been done previously for random forests \cite{ishwaran2007variable}, we defined the VIMP for a given feature $F$ as the difference between the prediction error for the original input matrix and the prediction error for the input matrix obtained by introducing noise to assignments of $F$. The continuous features were modified by adding noise drawn from $\mathcal{N}(0, \sigma \epsilon)$, where $\sigma$ is the standard deviation of $F$ and $\epsilon$ is a small constant, taken to be $0.1$ in our case. \\
For discrete features we flipped the features using the transformation: $x \rightarrow x*(1-s)+(1-x)*s$ where $s$ is drawn from a Bernoulli distribution. 
A high VIMP indicates a feature that has a strong effect on the survival.

\subsection{Evaluation on other datasets}
Finally, we assess all the comparative experiments on six other real-life datasets. As we can see in Table~\ref{table_datasets}, these datasets have different characteristics in terms of censored percentage, tied percentage, dimension (i.e., number of instances and number of features) and presence of missing values. These datasets are available on the Rdatasets github repository\footnote{\url{https://vincentarelbundock.github.io/Rdatasets/datasets.html}} accompanied with a complete description. Our method outperforms other models on most, but not all datasets. The results are summarized in Table \ref{c_index_results_other}. For each dataset, the best C-index value is in bold.

\begin{table}[!ht]
\centering
\small
\caption{
{\bf Characteristics of the real-life datasets.}}
\begin{tabular}{ |l|c|c|c|c|c| } 
\hline
Datasets & Nb. ($\%$) cens.& Nb. unique t ($\%$) & Nb. inst. & Nb. feat. & Missing val.\\
\hline
aids2 & 1082 (38.1) & 1013 (35.6) & 2843 & 12 & no \\
colon death & 477 (51.3) & 780 (84.0) & 929 & 37 & yes \\
colon recurrence & 461 (49.6) & 749 (80.6) & 929 & 37 & yes \\
flchain & 5705 (72.5) & 2977 (37.8) & 7874 & 23 & yes \\
mgus 2 tgt2 & 421 (30.4) & 272 (19.7) & 1384 & 5 & yes \\
nwtco & 3457 (85.8) & 2767 (68.7) & 4028 & 9 & no \\
\hline
\end{tabular}
\label{table_datasets}
\end{table}

\begin{table}[!ht]
\centering
\small
\caption{
{\bf C-index obtained for the different models tested on the other real-life datasets.}}
\label{cindex_values1}
\begin{tabular}{ |l|c|c|c| } 
\hline
Datasets & Cox Efron's & MLP Efron's & MLP Efron's + rank \\
\hline
aids2 & 0.5458 & 0.5495 & \textbf{0.5525}\\
colon death & 0.5024 & 0.5041 & \textbf{0.5603}\\
colon recurrence & 0.5456 & 0.6168 & \textbf{0.6170}\\
flchain & 0.7949 & 0.7974 & \textbf{0.8009}\\
mgus2 tgt2 & 0.6824 & 0.6884 & \textbf{0.6943}\\
nwtco & \textbf{0.7208} & 0.7092 & 0.7136\\
\hline
\end{tabular}
\label{c_index_results_other}
\end{table}

\section{Discussion and conclusion}
In conclusion, our method outperforms previous state-of-the-art methods in terms of the commonly used C-index metric. Moreover, it gives important clues about the survival prediction for different time thresholds. This shows the advantages of directly modeling the survival function.

\section{Note}
"The data reported here have been supplied by the Minneapolis Medical Research Foundation (MMRF) as the contractor for the Scientific Registry of Transplant Recipients (SRTR). The interpretation and reporting of these data are the responsibility of the author(s) and in no way should be seen as an official policy of or interpretation by the SRTR or the U.S. Government."

\section{Acknowledgments}
Special thanks to Adriana Romero for useful discussions.

\small
\bibliography{bibliography}
\bibliographystyle{plain}

\end{document}